\documentclass[conference]{IEEEtran}
\IEEEoverridecommandlockouts
\usepackage{cite}
\usepackage{amsmath,amssymb,amsfonts}
\usepackage{algorithmic}
\usepackage{graphicx}
\usepackage{textcomp}
\usepackage[table]{xcolor}
\usepackage{svg} 
\usepackage{amsmath} 
\usepackage{float}
\usepackage{multirow}
\usepackage{booktabs} 
\def\BibTeX{{\rm B\kern-.05em{\sc i\kern-.025em b}\kern-.08em
    T\kern-.1667em\lower.7ex\hbox{E}\kern-.125emX}}

\makeatletter
\newcommand{\linebreakand}{%
  \end{@IEEEauthorhalign}
  \hfill\mbox{}\par
  \mbox{}\hfill\begin{@IEEEauthorhalign}
}
\makeatother

\begin{document}

\title{Wearable ECG Quality Assessment: A
Deep Learning and Ambulatory
Context-Awareness Approach
\\

}

\author{\IEEEauthorblockN{Xiaopeng Mao}
\IEEEauthorblockA{\textit{Health Technology department} \\
\textit{Technical University of Denmark}\\
Kongens Lyngby, Denmark \\
s194408@dtu.dk}
\and
\IEEEauthorblockN{Marike Weisbjerg}
\IEEEauthorblockA{\textit{Health Technology department} \\
\textit{Technical University of Denmark}\\
Kongens Lyngby, Denmark \\
s194388@dtu.dk}
\and
\IEEEauthorblockN{Sadasivan Puthusserypady
}
\IEEEauthorblockA{\textit{Health Technology department} \\
\textit{Technical University of Denmark}\\
Kongens Lyngby, Denmark \\
sapu@dtu.dk}

}

\maketitle

\begin{abstract}
This paper presents and evaluates a Deep Learning-based (DL-based) Signal Quality Assessment (SQA) model to distinguish between clean and noisy ambulatory Electrocardiograms (ECG). The model is trained on Copenhagen Center for Health Technology-Contextualized Arrhythmia Database (CACHET-CADB), which, to the best of our knowledge, is the first ambulatory ECG database with both physical and patient-reported contextual data. The model shows stable performance on different databases such as MIT-databases and the latest PyhsioNet/Cinc Challenge 2021 databases. Subsequently, the paper demonstrates how complicated ECG noise can be investigated by the SQA model and the physical contextual data.   
\end{abstract}

\begin{IEEEkeywords}
Deep learning, ECG, SQA, contextualized analysis
\end{IEEEkeywords}

\section{Introduction}
As a major heart disease, arrhythmias require continuous monitoring with portable ECG device. Hence, it is important to prevent noise from obscuring the ECG. Under free-living conditions, the noise level is particularly severe compared to in-hospital recording. To assess the quality of ambulatory ECG, this study provides a DL SQA model trained on a contextualized ambulatory ECG arrhythmia data set, CACHET-CADB~\cite{CACHET}. The strength and weakness of the model is evaluated on the well-known public databases such as MIT-BIH-Arrhythmia and the recent databases from PhysioNet/Cinc challenge 2021. The results from the contextualized analysis are also included to show how the aforementioned model combined with the physical ECG context can provide a deeper insight of ECG noise and thus improve the recording environment. In summary, this study aims to evaluate a state-of-the-art DL SQA model on multiple databases and to explore the new possibilities of a novel contextualized ECG database.

\section{Methods}

\subsection{Datasets}
CACHET-CADB consists of 259 days of continuous ambulatory ECG recorded from 24 people, where 21 are Atrial Fibrillation (AFIB) patients while the rest are healthy individuals. The signal is sampled in 1,024 Hz from one single precordial lead as described in~\cite{CACHET}. 1,602 of 10 s ECGs are annotated by two cardiologists independently. The labelled classes are Normal Sinus Rhythm (NSR), AFIB, noise and others. Table 1 provides an overview of the labelled data.
\begin{table}[H]
\begin{footnotesize}
\caption{ECG rhythms from CACHET-CADB.}
\label{tab:labelled}
\begin{center}
\begin{tabular}{ccccc}
    \toprule
\textbf{Category} & \textbf{NSR} & \textbf{AFIB} & \textbf{Other} & \textbf{Noise} \\\hline\\[-1.5ex]
Number of signals & 615                                & 747                               & 19              & 221   \\ 
Considered class & Clean & Clean & Clean & Noisy \\\hline
    \bottomrule
\end{tabular}
\end{center}
\end{footnotesize}
\end{table} 
In this study, the aim of the SQA is to separate ECG noise from anything else. Hence, an ECG that is either NSR, AFIB or other rhythms will be identified as "clean", otherwise the ECG is "noisy". The labelled ECGs are treated as the test set to evaluate the DL model.

\subsection{Continuous Wavelet Transformation}
\begin{figure}[H]
    \hspace{-0.5cm}
    \includegraphics[width=90mm]{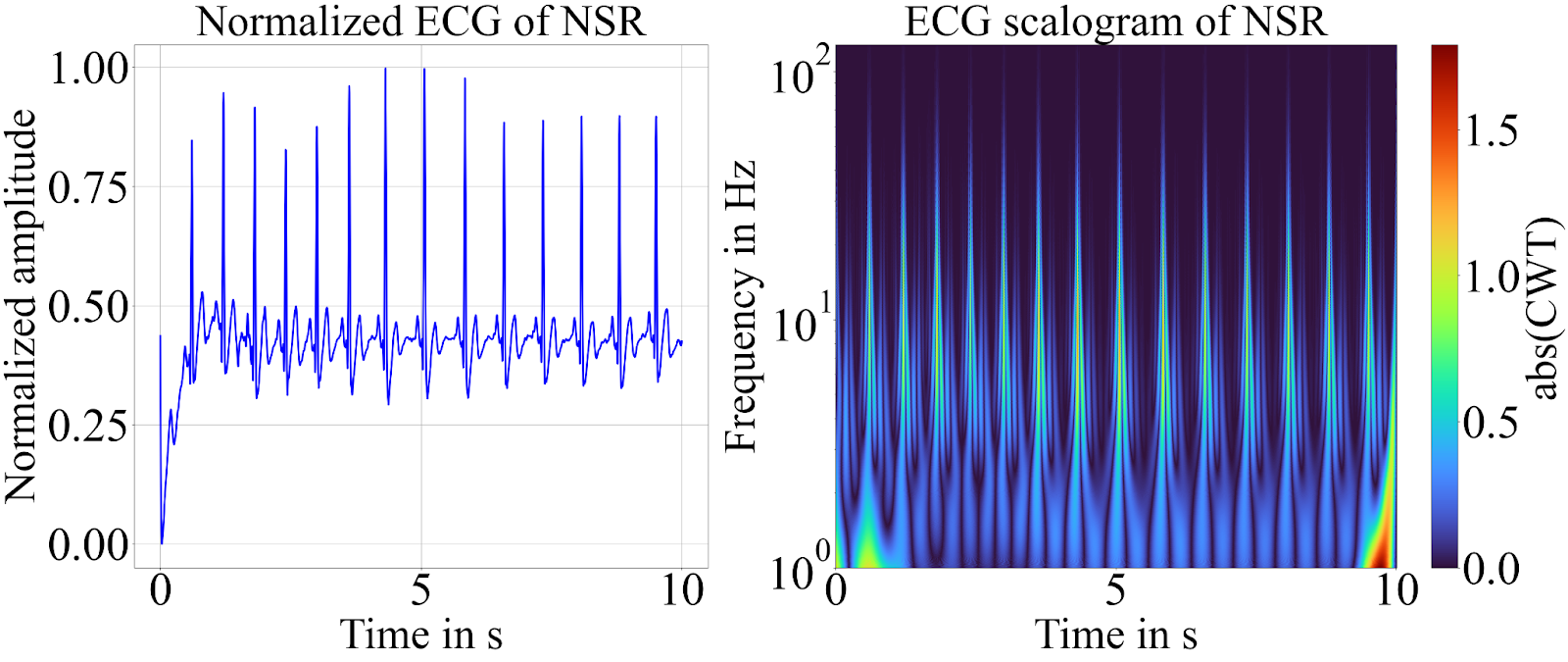}
    \caption{Scalogram of a 10 s NSR}
    \label{IM:scalogram_1}
\end{figure}

Continuous Wavelet Transformation (CWT) transforms the 1D ECGs into 2D scalograms. It is defined as the absolute values of the CWT-coefficients~\cite{scalogram}. The advantage of CWT is that it provides a time-frequency representation of the signal, whereas plain ECG only provides a time domain representation. Hence, image classification between clean and noisy ECGs can be performed on scalograms. The CWT-coefficients are calculated by the formula in (\ref{eq:CWT}).
\begin{equation}
    C_{\Psi,x}(a,b) = \frac{1}{\sqrt{|a|}}\cdot \int_{-\infty}^{\infty}x(t)\cdot \Psi^*\left(\frac{t-b}{a}\right)dt,\, a\neq 0,
    \label{eq:CWT}
\end{equation}
where $x(t)$ is the input ECG, $\Psi*$ is time-scaled and time-shifted wavelet function of the mother wave, $\Psi$, $a$ is the time-scaling parameter and $b$ is the time-shifting parameter~\cite{CWT_formula}.
The frequency is defined in equation (\ref{eq:freq}).
\begin{equation}
    f=\frac{f_c \cdot f_s}{a},\, a\neq 0,
    \label{eq:freq}
\end{equation}
where $f_s$ denotes the ECG sampling frequncy, i.e. 1,024 Hz, $f_c$ denotes the central frequency of $\Psi$~\cite{CWT_frequency}.
In this study, the frequency is in a logarithmic scale since it better represents the scalogram. Figure \ref{IM:scalogram_1} and 2 show a scalogram of a 10 s NSR and noise ECG, respectively.

\begin{figure}[H]
    \hspace{-0.5cm}
    \includegraphics[width=90mm]{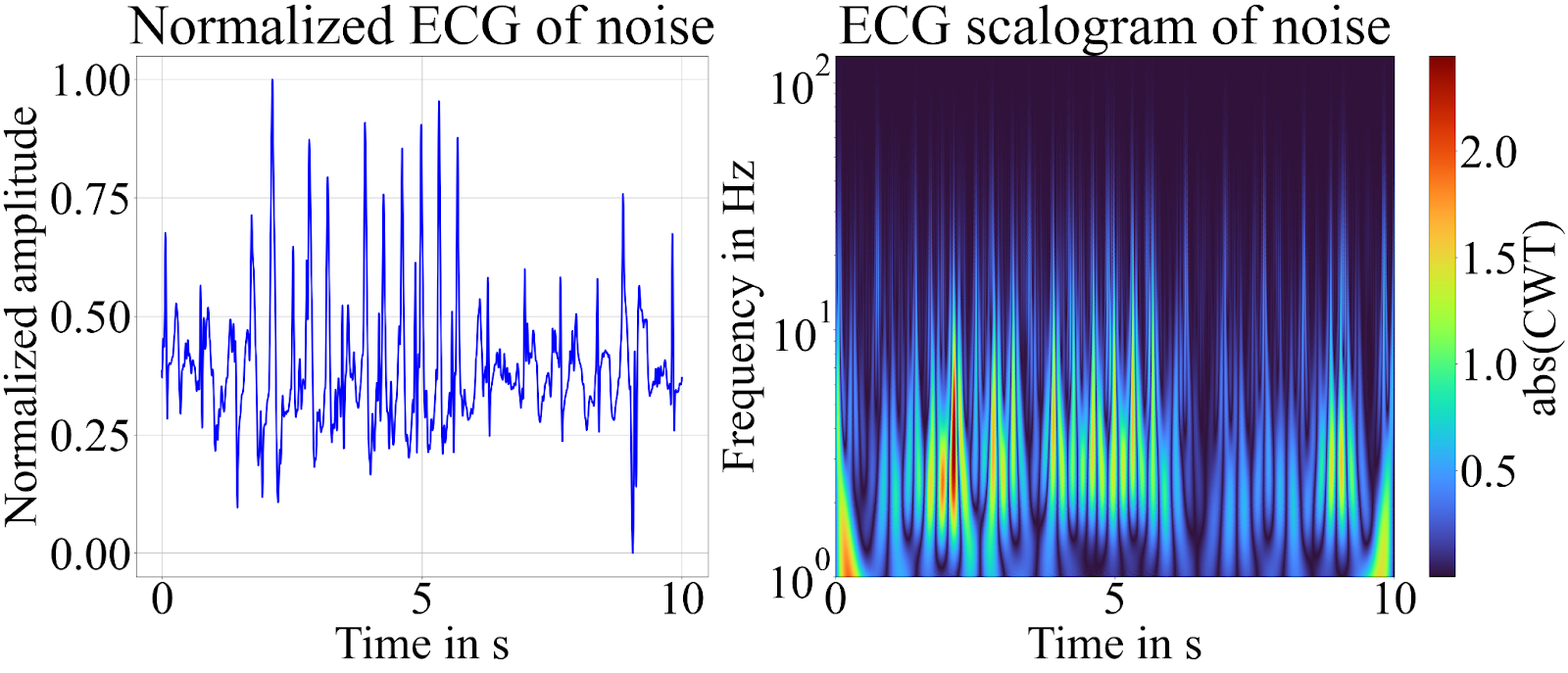}
    \caption{Scalogram of a 10 s noise}
    \label{IM:scalogram_2}
\end{figure}

CWT is widely applied to public databases such as MIT-BIH arrhythmia and FysioNet/CinC Challenge to train binary/multi-class ECG SQA~\cite{CWT_1, CWT_2}. In~\cite{CWT_2}. Huerta \textit{et al.} used Transfer Learning (TL) to train a series of DL models. In this work, InceptionV3 is used to classify the ECGs.

\subsection{Model architecture}
\begin{figure}[H]
\hspace{0.5cm}
\centerline{\includegraphics[width=90mm]{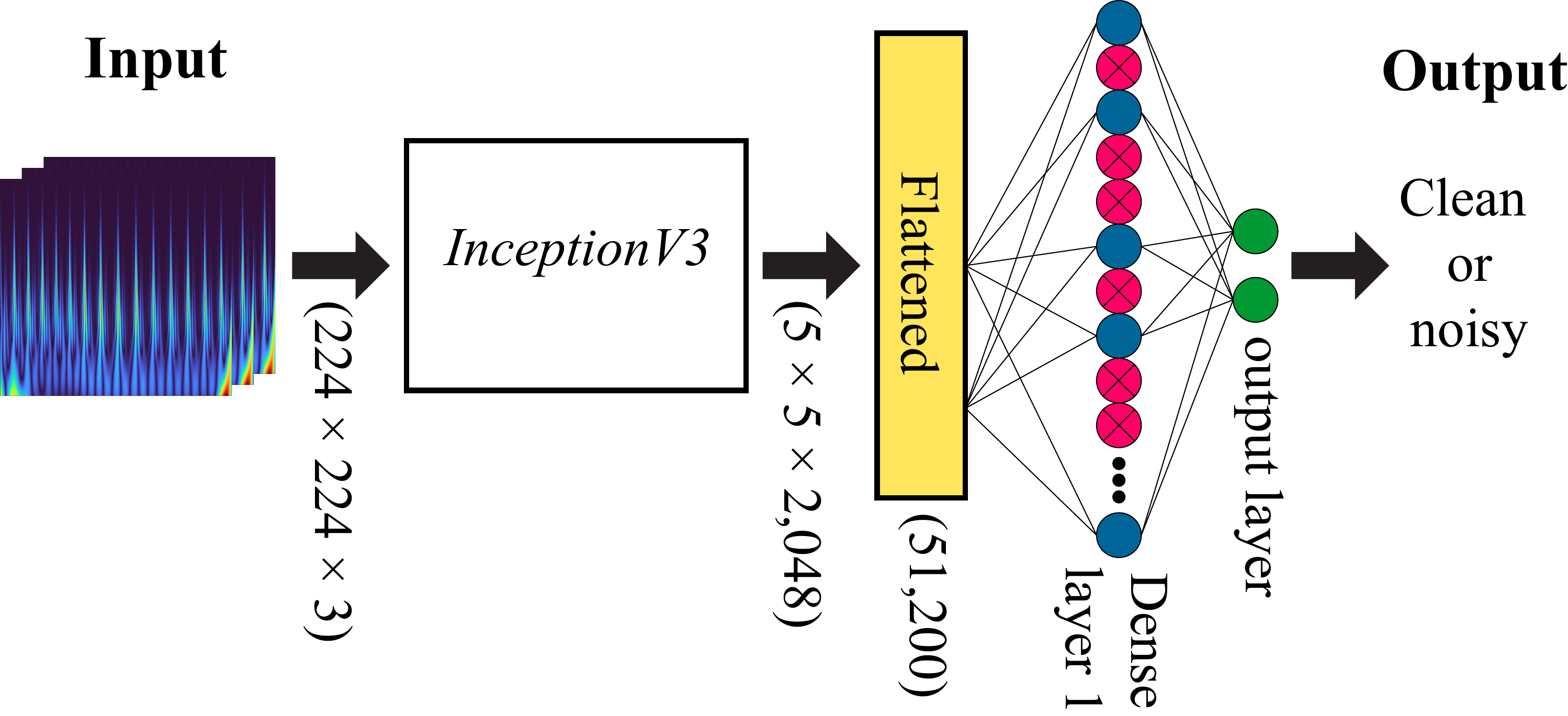}}
\caption{Architecture of the final model. The three input images are identical. The red neurons in Dense layer 1 are prohibited by a dropout probability.}
\label{IM:model}
\end{figure}

The input dimensions are 224 $\times$ 224 $\times$ 3. The input goes through the InceptionV3 modules to obtain the new output dimensions of 5 $\times$ 5 $\times$ 2048, which is flattened before putting into the dense layers. During the process, all parameters of InceptionV3 are frozen to prevent overfitting. A dropout layer is also added to Dense layer 1 to stabilize the network training. Figure \ref{IM:model} illustrates the above-mentioned model achitecture.
The hyper-parameters are tuned by using random search. Table \ref{tab:hyperpara} shows the final values for the hyper-parameters of the model.

\begin{table}[H]
\caption{Optimized hyper-parameters of the DL model.}
\fontsize{9}{9}\selectfont
\begin{center}
\begin{tabular}{cc}
\toprule
\textbf{Hyper-parameter} & \textbf{Value} \\\hline\\[-1.5ex]
Dense layer 1 size      & 20             \\
Learning rate           & $1\cdot10^{-5}$           \\
Dropout probability     & 0.6            \\
Batch size              & 20             \\
Epochs                  & 40             \\ \hline
\bottomrule
\end{tabular}
\end{center}
\label{tab:hyperpara}
\end{table}

\subsection{Contextualized analysis}
The contextualized analysis is done by using the physical parameters that are automatically measured by the barometer and accelerometer of the ECG device~\cite{CACHET}. The parameters include body positions and activity classes listed in Table \ref{context_parameter}.

\begin{table}[H]
 \caption{The contextualized parameters presented in~\cite{CACHET}.}
\label{context_parameter}
\fontsize{9}{9}\selectfont
\begin{center}
    \begin{tabular}{ l l }
    \toprule
    \textbf{Attribute}     & \textbf{Parameter}                        \vspace{0.2cm} \\\hline\\[-1.5ex]
    & 0: Unknown \\
    & 1: Lying supine \\
    & 2: Lying left \\ 
    & 3: Lying prone \\
Body position & 4: Lying right \\
    & 5: Upright \\
    & 6: Sitting/lying \\
    & 7: Standing \\
    & 99: Not worn \\
 \vspace{-1mm} \\\hline\\[-1.5ex]

    & 0: Unknown \\
    & 1: Lying  \\
    & 2: Sitting/standing \\ 
    & 3: Cycling \\
    & 4: Slope up \\
    & 5: Jogging \\
Activity class   & 6: Slope down \\
    & 7: Walking \\
    & 8: Sitting/lying \\
    & 9: Standing \\
    & 10: Sitting/lying/standing\\
    & 11: Sitting\\
    & 99: Not worn \\     \vspace{-1mm} \\\hline
    \bottomrule
    \end{tabular}
\end{center}
\end{table}

\begin{figure}[H]
\centerline{\includegraphics[width=90mm]{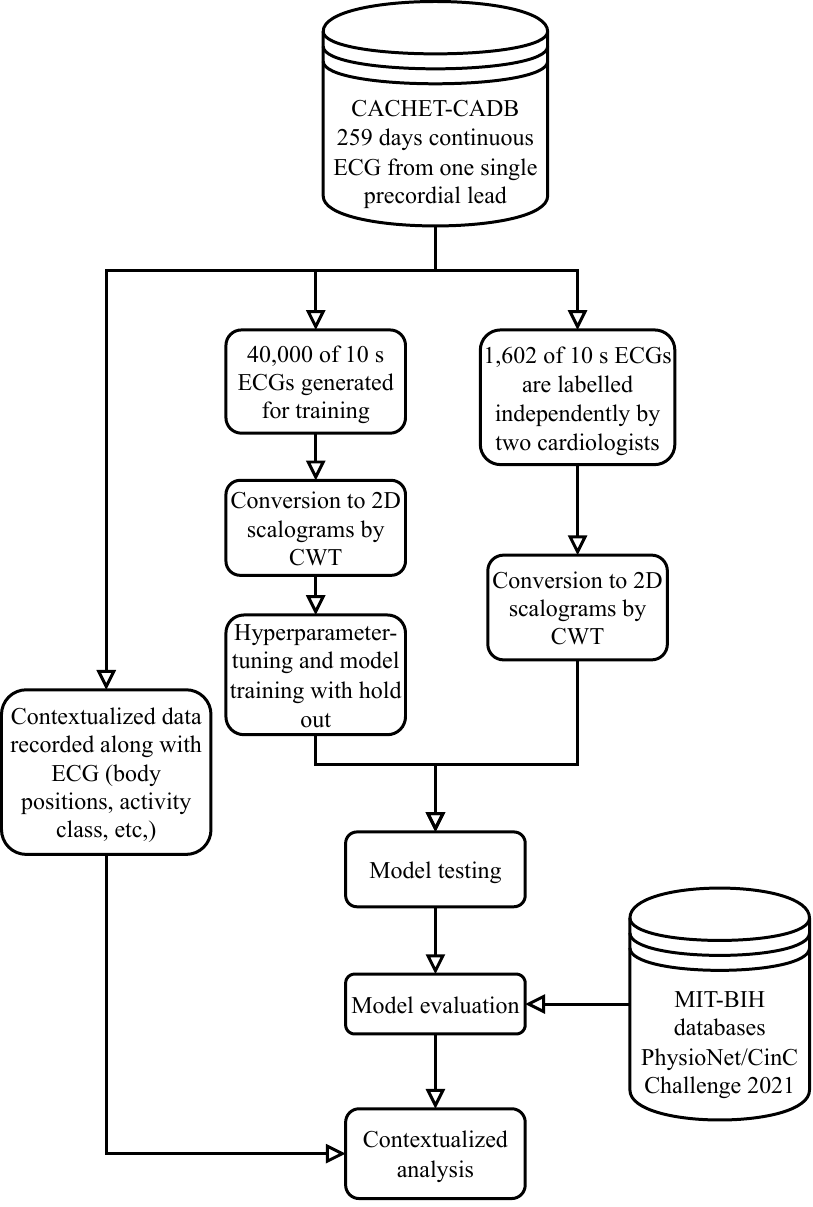}}
\caption{Flow diagram of the work.}
\label{IM:flow_diagram}
\end{figure}

As the standard pre-processing, all ECGs are bandpass-filtered within 0.5 and 50 Hz and smoothed by using Savitzky-Golay filter \cite{CACHET, Bioelectrical_signal_pro}. The signals are then re-sampled to 256 Hz. 20,000 clean and 20,000 noisy ECGs of 10 s are generated from CACHET-CADB by using a set of decision rules combined with template matching from~\cite{PPG}. The method is adjusted to optimize its performance on AFIB. Figure \ref{IM:flow_diagram} shows the workflow throughout the entire study. 

As for the network training, Adaptive Moment Estimation (Adam) optimizer is used~\cite{CWT_1}. The cross entropy function is used as the loss function for the DL model~\cite{loss_func}.
The software used for training is Python and TensorFlow.
The model is trained by using 3 nodes of 4 $\times$ Tesla V100 (32 GB) with NVlink (owned by DTU
Compute).

\section{Results}

\subsection{Performance on CACHET-CADB test set}\label{AA}
Table \ref{tab:CACHET} shows the confusion matrix of the DL SQA model.

\begin{table}[H]
\centering
\caption{Confusion matrix of the model performance on CACHET-CADB test set.}
\begin{tabular}{cc|cc|c}
\cline{1-4}
\multicolumn{2}{|c|}{}                                                                                             & \multicolumn{2}{c|}{}                                                                      &                                                                                               \\
\multicolumn{2}{|c|}{}                                                                                             & \multicolumn{2}{c|}{\multirow{-2}{*}{Actual class}}                                        & \multicolumn{1}{l}{}                                                                          \\ \cline{3-4}
\multicolumn{2}{|c|}{}                                                                                             & \multicolumn{1}{c|}{}                                 &                                    &                                                                                               \\
\multicolumn{2}{|c|}{\multirow{-4}{*}{\begin{tabular}[c]{@{}c@{}}1,602 labelled\\ signals in total.\end{tabular}}} & \multicolumn{1}{c|}{\multirow{-2}{*}{True}}           & \multirow{-2}{*}{False}            & \multicolumn{1}{l}{}                                                                          \\ \hline
\multicolumn{1}{|c|}{}                                                 & True                                      & \multicolumn{1}{c|}{\cellcolor[HTML]{bee881}TP = 215} & \cellcolor[HTML]{ffe881}FP = 128   & \multicolumn{1}{c|}{\begin{tabular}[c]{@{}c@{}}Predicted \\ positives\\ = 343\end{tabular}}   \\ \cline{2-5} 
\multicolumn{1}{|c|}{\multirow{-4}{*}{Prediction}}                     & False                                     & \multicolumn{1}{c|}{\cellcolor[HTML]{ffe881}FN = 6}   & \cellcolor[HTML]{bee881}TN = 1,253 & \multicolumn{1}{c|}{\begin{tabular}[c]{@{}c@{}}Predicted \\ negatives\\ = 1,259\end{tabular}} \\ \hline
                                                                       &                                           & \multicolumn{1}{c|}{}                                 &                                    & \multicolumn{1}{c|}{}                                                                         \\
\multicolumn{1}{l}{}                                                   & \multicolumn{1}{l|}{}                     & \multicolumn{1}{c|}{\multirow{-2}{*}{Se = 97.3 \%}}   & \multirow{-2}{*}{Sp = 90.7 \%}     & \multicolumn{1}{c|}{\multirow{-2}{*}{Acc = 91.6\%}}                                           \\ \cline{3-5} 
\end{tabular}
\label{tab:CACHET}
\end{table}

\begin{figure}[H]
\centerline{\includegraphics[width=90mm]{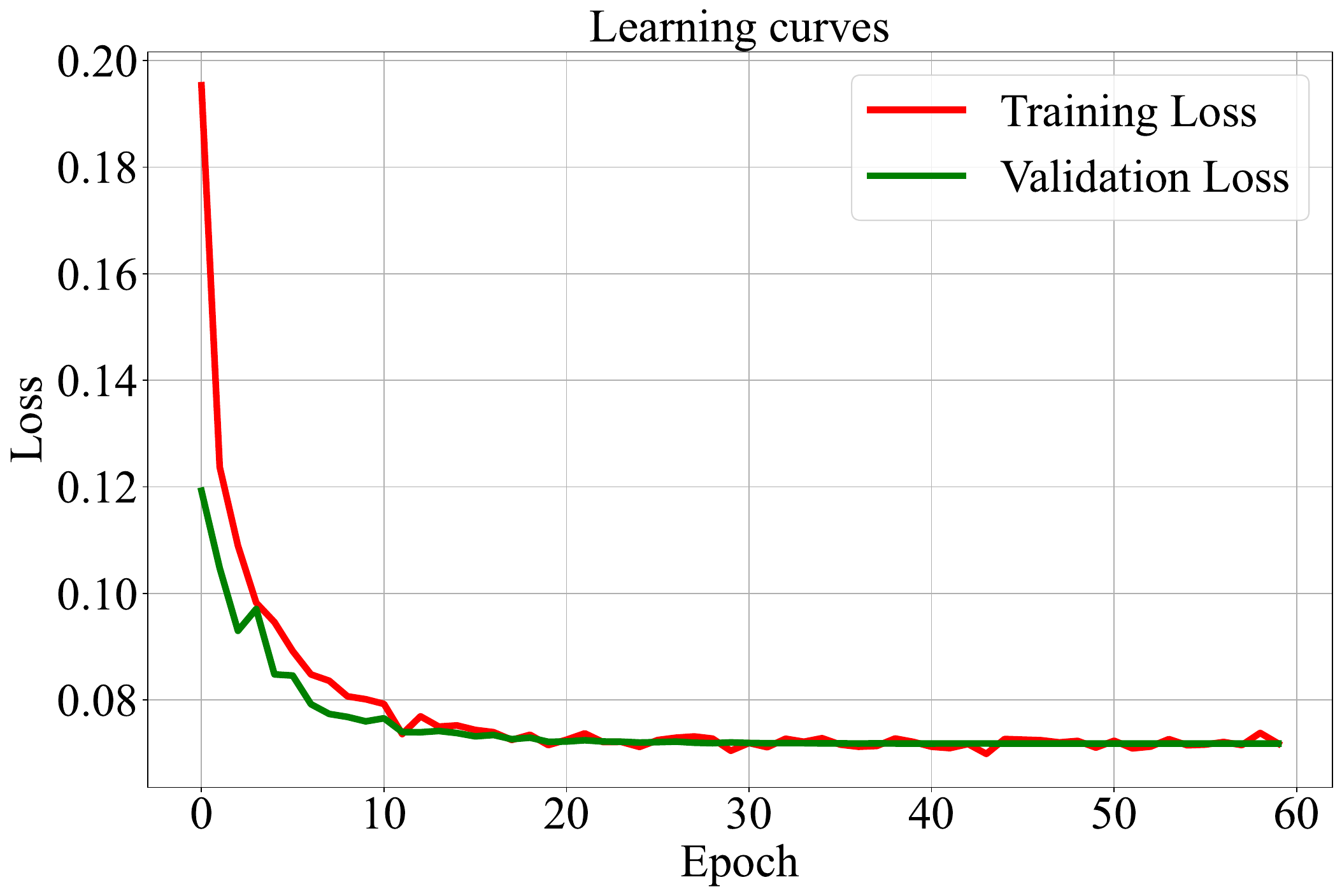}}
\caption{The learning curves of the model.}
\label{IM:lr_curves}
\end{figure}

In Table \ref{tab:CACHET}, it can be seen that the accuracy lies closer to the specificity than sensitivity due to the fact that the number of noisy signals are significantly lower than the number of clean signals.
The learning curves of the model are shown in Figure \ref{IM:lr_curves}. The training set consists of 32,000 images while the validation set consists of 8,000 images. The clean and noisy ECGs are evenly distributed among them. 

The convergence of both loss curves in Figure \ref{IM:lr_curves} indicates a stable learning process with the optimized hyper-parameters from Table \ref{tab:hyperpara}. An exponentially decaying learning schedule with a base of 0.18 and a step of 10 is used, so the learning rate decreases with 82 \% after every 10 epochs.

\subsection{Performance on MIT-BIH-NSR, MIT-BIH-Arrhythmia and MIT-BIH-Noise Stress Test Database}\label{sec:MIT}

3 records are randomly chosen from the MIT-BIH-NSR database. The entire arrhythmia database is used. 3 pure noise signals are used from MIT-BIH-Noise Stress Test database (the noise types include baseline wander, muscle artifact and electrode motion artifact). Lead 2 ECG is used for the model evaluation because this particular lead most clearly expresses the QRS complex~\cite{Bioelectrical_signal_pro}.
\begin{table}[H]
 \caption{Prediction on different MIT-BIH databases.}
\hspace{-0.5cm}
\begin{footnotesize}
\begin{tabular}{cccc}
\toprule
\textbf{Database}                                                   & \textbf{Record amount} & \textbf{Duration (h)} & \textbf{\begin{tabular}[c]{@{}c@{}}Percentage of \\ predicted clean\\ signals (\%)\end{tabular}} \\\hline\\[-1.5ex]
MIT-BIH-NSR                                                         & 3                      & 48                    & 98.2                                                                                             \\[1ex]
MIT-BIH-Arrhythmia                                                  & 48                     & 24                    & 76.9                                                                                             \\[1ex]
\begin{tabular}[c]{@{}c@{}}MIT-BIH-Noise\\ Stress Test\end{tabular} & 3                      & 1.5                   & 2.2                                                                                              \\ \hline
\bottomrule
\end{tabular}
\end{footnotesize}
\label{tab:MIT}
\end{table}

The individual record of MIT-BIH-Arrhythmia shows that the model tends to fail on rhythms with a rather abnormal morphology such as paced beats combined with Premature Ventricular Contraction (PVC).

\subsection{Performance on PhysioNet/CinC challenge 2021}
The Chapman-Shaoxing database is one of the most recent databases from the Chapman University and Shaoxing People's Hospital~\cite{Chapman}. The database contains 10,646 (10,247 available from the challenge) of 10 s clean ECGs from 10,646 heart patients. The sampling frequency is 500 Hz. Table \ref{tab:cinc_2021} shows the model performance categorized for each ECG rhythm. 
\begin{table}[H]
    \caption{DL model performance on the Chapman-Shaoxing database. The worst performances are marked with orange.}
\begin{footnotesize}
\hspace{-0.5cm}
\begin{tabular}{cccc}
\toprule
\textbf{Category}                                                                                & \textbf{\begin{tabular}[c]{@{}c@{}}Number of\\ signals\end{tabular}} & \textbf{\begin{tabular}[c]{@{}c@{}}Number of \\ predicted clean\\ signals\end{tabular}} & \textbf{\begin{tabular}[c]{@{}c@{}}Percentage of\\ predicted clean\\ signals (\%)\end{tabular}} \\\hline\\[-1.5ex]
\begin{tabular}[c]{@{}c@{}}Sinus Bradycardia\\ (SB)\end{tabular}                                & 3,889                                                                & 3,605                                                                                   & 92.7                                                                                            \\[3ex]
\begin{tabular}[c]{@{}c@{}}Normal Sinus\\ Rhythm\\ (NSR/SR)\end{tabular}                          & 1,826                                                                & 1,661                                                                                   & 91.0                                                                                            \\[3ex]
\rowcolor[HTML]{FFCC67} 
\begin{tabular}[c]{@{}c@{}}Atrial Fibrillation\\ (AFIB)\end{tabular}                            & 1,780                                                                & 1,171                                                                                   & 65.8                                                                                            \\[3ex]
\begin{tabular}[c]{@{}c@{}}Sinus Tachycardia\\ (ST)\end{tabular}                                & 1,568                                                                & 1,305                                                                                   & 83.2                                                                                            \\[2ex]
\rowcolor[HTML]{FFCC67} 
\begin{tabular}[c]{@{}c@{}}Supraventricular \\ Tachycardia\\ (SVT)\end{tabular}                 & 587                                                                  & 345                                                                                     & 58.8                                                                                            \\[4ex]
\rowcolor[HTML]{FFCC67} 
\begin{tabular}[c]{@{}c@{}}Atrial Flutter\\ (AF)\end{tabular}                                   & 445                                                                  & 280                                                                                     & 62.9                                                                                            \\[3ex]
\rowcolor[HTML]{FFCC67} 
\begin{tabular}[c]{@{}c@{}}Atrial Tachycardia\\ (AT)\end{tabular}                               & 121                                                                  & 73                                                                                      & 60.3                                                                                            \\[3ex]
\begin{tabular}[c]{@{}c@{}}Atrioventricular Node\\ Reentrant Tachycardia\\ (AVNRT)\end{tabular} & 16                                                                   & 14                                                                                      & 87.5                                                                                            \\[4ex]
\begin{tabular}[c]{@{}c@{}}Atrioventricular \\ Reentrant Tachycardia\\ (AVRT)\end{tabular}      & 8                                                                    & 7                                                                                       & 87.5                                                                                            \\[4ex]
\begin{tabular}[c]{@{}c@{}}Sinus Atrium to\\ Atrial Wandering \\ Rhythm\\ (SAAWR)\end{tabular}  & 7                                                                    & 7                                                                                       & 100                                                                                             \\[5ex]
Total                                                                                           & 10.247                                                               & 8.468                                                                                   & 82.6                                                                                            \\ \hline
\bottomrule
\end{tabular}
\end{footnotesize}
\label{tab:cinc_2021}
\end{table}

The Ningbo database is the largest database from the 2021 challenge. The database contains 40,258 ECGs (34,905 available from the challenge).
The ECGs are all from different patients, 10 s long, and sampled with 500 Hz. All annotations are manually identified, so the model performance are evaluated for all ECG types (see Table \ref{tab:cinc_2021_2}).

\begin{table}[H]
    \caption{DL model performance on the Ningbo arrhythmia database. The worst performances are marked with orange.}
\begin{footnotesize}
\begin{tabular}{cccc}
\toprule
\textbf{Category*}                                                                               & \textbf{\begin{tabular}[c]{@{}c@{}}Number of \\ signals\end{tabular}} & \textbf{\begin{tabular}[c]{@{}c@{}}Number of \\ predicted clean\\ signals\end{tabular}} & \textbf{\begin{tabular}[c]{@{}c@{}}Percentage of\\ predicted clean\\ signals (\%)\end{tabular}} \\\hline\\[-1.5ex]
\begin{tabular}[c]{@{}c@{}}Sinus Bradycardia\\ (SB)\end{tabular}                                & 11,919                                                                 & 11,059                                                                                  & 92.8                                                                                            \\[3ex]
\begin{tabular}[c]{@{}c@{}}Normal Sinus \\ Rhythm\\ (NSR/SR)\end{tabular}                       & 6,058                                                                  & 5,489                                                                                   & 90.6                                                                                            \\[3ex]
\rowcolor[HTML]{FFCC67} 
\begin{tabular}[c]{@{}c@{}}Atrial Flutter\\ (AF)\end{tabular}                                   & 5,164                                                                  & 3,498                                                                                   & 67.7                                                                                            \\[2ex]
\begin{tabular}[c]{@{}c@{}}Sinus Tachycardia\\ (ST)\end{tabular}                                & 3,820                                                                  & 3,182                                                                                   & 83.3                                                                                            \\[2ex]
Sinus arrhythmia                                                                                & 1,294                                                                  & 1,180                                                                                   & 91.2                                                                                            \\[2ex]
\rowcolor[HTML]{FFCC67} 
\begin{tabular}[c]{@{}c@{}}Premature Atrial \\ Contraction\\ (PAC)\end{tabular}                 & 975                                                                    & 681                                                                                     & 69.8                                                                                            \\[3ex]
T wave abnormal                                                                                 & 896                                                                    & 702                                                                                     & 78.3                                                                                            \\[2ex]
\rowcolor[HTML]{FFCC67} 
\begin{tabular}[c]{@{}c@{}}ST interval \\ abnormal\end{tabular}                                 & 401                                                                    & 268                                                                                     & 66.8                                                                                            \\[2ex]
\begin{tabular}[c]{@{}c@{}}Left ventricular \\ hypertrophy\end{tabular}                         & 390                                                                    & 286                                                                                     & 73.3                                                                                            \\[3ex]
\rowcolor[HTML]{FFCC67} 
\begin{tabular}[c]{@{}c@{}}Non-specific \\ intraventricular \\ conduction delay\end{tabular}    & 364                                                                    & 187                                                                                     & 51.4                                                                                            \\[4ex]
\rowcolor[HTML]{FFCC67} 
ST segment changes                                                                              & 331                                                                    & 221                                                                                     & 66.8                                                                                            \\[2ex]
\begin{tabular}[c]{@{}c@{}}Right axis \\ deviation\end{tabular}                                 & 301                                                                    & 218                                                                                     & 72.4                                                                                            \\[2ex]
\rowcolor[HTML]{FFCC67} 
ST Depression                                                                                   & 275                                                                    & 180                                                                                     & 65.5                                                                                            \\[1ex]
\rowcolor[HTML]{FFCC67} 
Paced beats                                                                                     & 265                                                                    & 153                                                                                     & 57.7                                                                                            \\[1ex]
\rowcolor[HTML]{FFCC67} 
\begin{tabular}[c]{@{}c@{}}First degree \\ atrioventricular \\ block\end{tabular}               & 233                                                                    & 162                                                                                     & 69.5                                                                                            \\[4ex]
\rowcolor[HTML]{FFCC67} 
\begin{tabular}[c]{@{}c@{}}Complete right\\ bundle branch \\ block\end{tabular}                 & 225                                                                    & 133                                                                                     & 59.1                                                                                            \\[4ex]
\rowcolor[HTML]{FFCC67} 
Inverted T wave                                                                                 & 119                                                                    & 70                                                                                      & 58.8                                                                                            \\[1ex]
Atrial rhythm                                                                                   & 112                                                                    & 98                                                                                      & 87.5                                                                                            \\[1ex]
\begin{tabular}[c]{@{}c@{}}Prolonged \\ QT interval\end{tabular}                                & 110                                                                    & 83                                                                                      & 75.5                                                                                            \\[2ex]
\rowcolor[HTML]{FFCC67} 
R wave                                                                                          & 108                                                                    & 63                                                                                      & 58.3                                                                                            \\[1ex]
\begin{tabular}[c]{@{}c@{}}Counterclockwise \\ vectorcardiographic \\ loop\end{tabular}         & 100                                                                    & 85                                                                                      & 85.0                                                                                            \\[3ex]
\rowcolor[HTML]{FFCC67} 
\begin{tabular}[c]{@{}c@{}}59 other abnormal \\ ECG rhythms\\ (each less than 100)\end{tabular} & 1,445                                                                  & 740                                                                                     & 51.2                                                                                            \\[3ex]
Total                                                                                           & 34,905                                                                 & 28,738                                                                                  & 82.3                                                                                            \\ \hline
\bottomrule
\end{tabular}
\end{footnotesize}
\\[1ex]
\text{*Manually identified by SNOMED-CT code look up.}
\label{tab:cinc_2021_2}
\end{table}

In summary, the evaluations show that the model is prone to make error when the morphology of the ECG deviates significantly from the normal sinus wave while a change of beat frequency (e.g. sinus bradycardia or tachycardia) does not affect the model performance severely.

\subsection{Contextualized analysis}
\begin{figure}[H]
\centerline{\includegraphics[width=85mm]{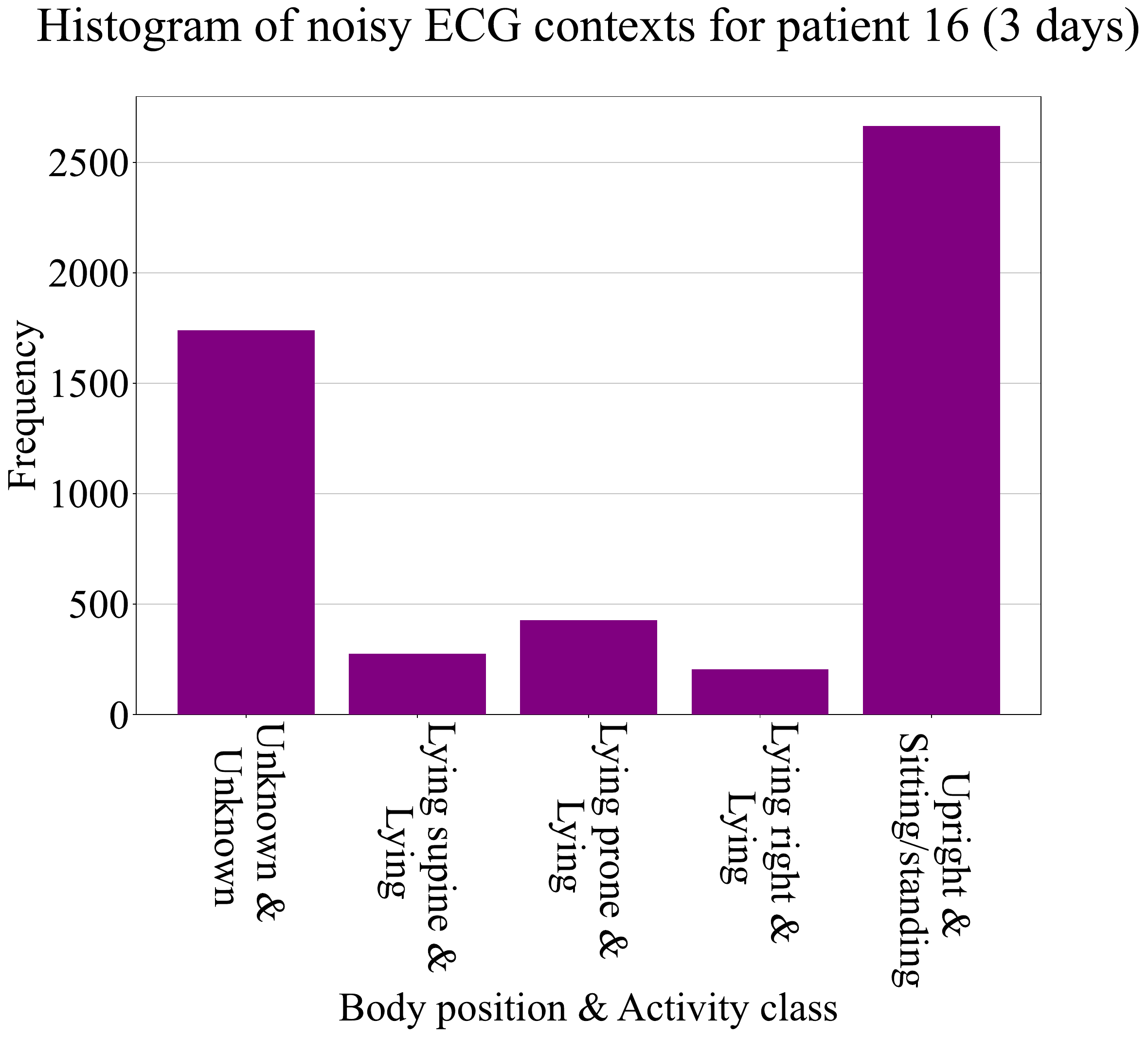}}
\caption{The underlying physical contexts of the noisy signals.}
\label{IM:histogram_noisy}
\end{figure}

\begin{figure}[H]
\centerline{\includegraphics[width=85mm]{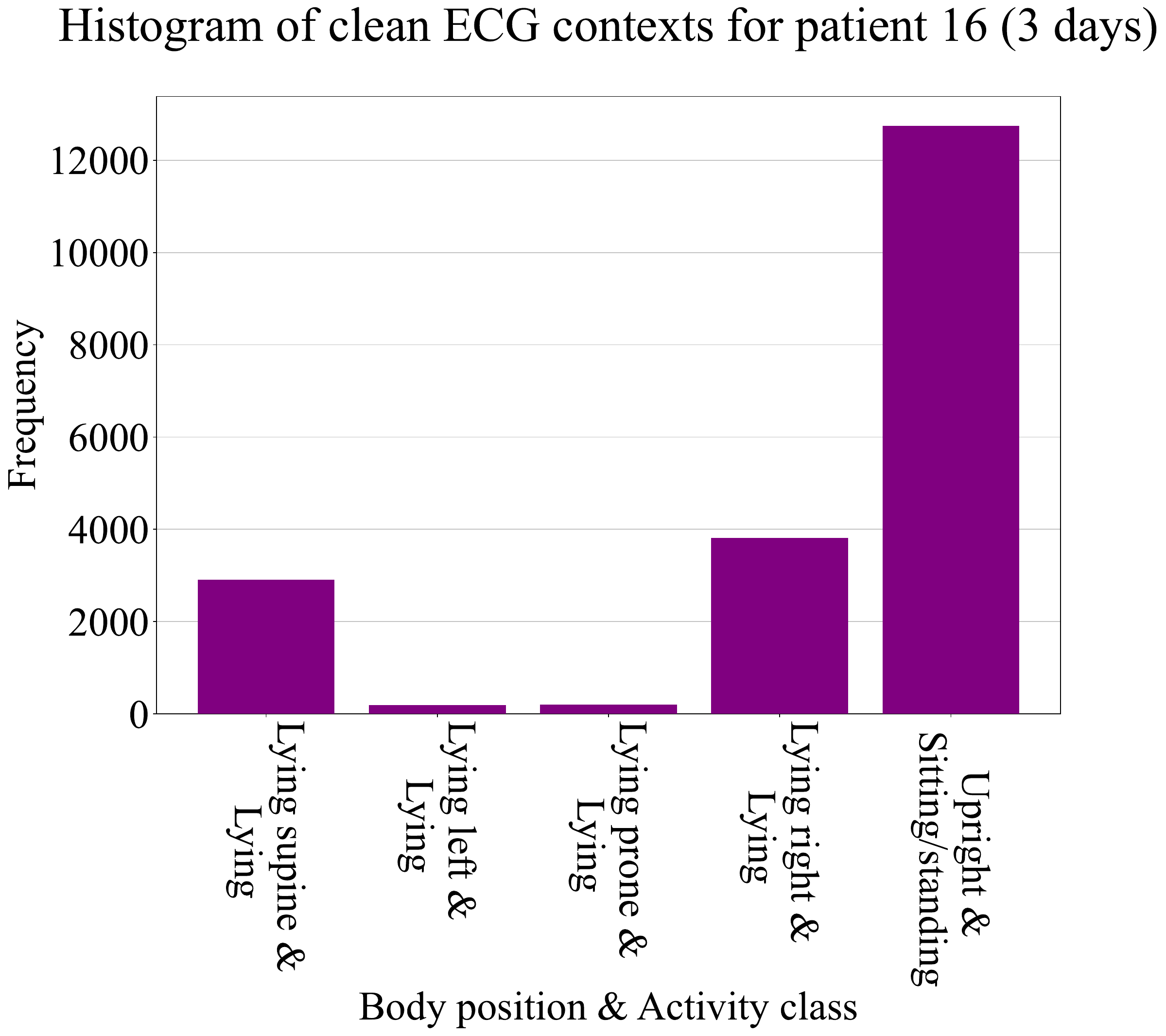}}
\caption{The underlying physical contexts of the clean signals.}
\label{IM:histogram_clean}
\end{figure}

The histograms in Figure \ref{IM:histogram_noisy} and \ref{IM:histogram_clean} show the combinations of a body position and an activity class that are present when the recorded ECG is noisy or clean, respectively.

In Figure \ref{IM:histogram_noisy}, it can be seen that the ECG noise is mostly present for sitting/standing in an upright body position and a combination of unknowns. As for the clean ECGs, sitting/standing upright is also present as the most frequent compared to the other combinations. This indicates that the overall signal quality might be improved when the unknown body position and activity class are avoided. A more precis identification of the unknown parameters can be made by backtracking their time periods and by inquiring the patient about the underlying events for the unknown parameters.

\section{Discussion}

In Table \ref{tab:cinc_2021}, it can be seen that the predictions on the sinus waves are particularly accurate. On the other hand, the model struggles to make correct predictions on the abnormal atrial waves such as AFIB.

The results from Table \ref{tab:cinc_2021_2} show that the abnormal ST segments, ST intervals and cardiac blocks are particularly difficult for the model to predict. This makes sense since the CACHET-CADB does not contain any patient, who suffers a disease that induces these rhythms. The predictions are poor especially on the rhythms with abnormal PQRST features. This agrees with the result from MIT-BIH-Arrhythmia in Table \ref{tab:MIT}.

In conclusion, the ECG rhythms with normal PQRST character are more likely to be predicted correctly by the model, whereas other pathological waveforms (the orange ones in Table \ref{tab:cinc_2021} and \ref{tab:cinc_2021_2}) with different morphologies are more likely to lead to wrong predictions. 

As for the contextualized analysis, it is important to notice that the clean and noisy ECGs are not reviewed by any cardiologist. Hence, it is necessary to be cautious about the analysis results presented in Figures \ref{IM:histogram_noisy} and \ref{IM:histogram_clean}.

Nevertheless, the identification of the physical context with respect to ECG signal quality will mark a small beginning within the huge area of signal context research.

\section*{Acknowledgment}

This project was conducted in summer 2022 as part of a bachelor project. The project was accomplished at the Department of Health Technology and Copenhagen Center for Health Technology (CACHET) at the Technical University of Denmark (DTU). The computational resources were provided by DTU Compute.

\end{document}